\documentclass[11pt]{article} 
\usepackage[paperheight=9.44in,paperwidth=6.69in,left=1.85cm,right=1.8cm,bottom=2.5cm,top=3cm,twoside]{geometry}

\usepackage{subfig} 

\usepackage[ruled,vlined,boxed]{algorithm2e}
\usepackage{authblk} 
\usepackage{fancyhdr} 
\usepackage{lastpage} 
\usepackage{textcomp} 
\usepackage{amsmath} 
\usepackage{amssymb} 
\usepackage{amsthm} 
\usepackage{graphicx} 
\usepackage{multirow} 
\usepackage{hyperref} 
\usepackage{longtable} 
\usepackage{float} 
\usepackage{enumitem} 
\usepackage[T1]{fontenc} 
\usepackage[utf8]{inputenc} 

\newcommand{\linia}{\noindent\rule{\linewidth}{0.25mm}\hrulefill} 

\SetAlFnt{\small}
\SetAlCapFnt{\small}
\SetAlCapNameFnt{\small}
\SetAlCapHSkip{0pt}
\IncMargin{-\parindent}

\usepackage{times}
\usepackage{titlesec}

\titleformat*{\section}{\large\bfseries}
\titleformat*{\subsection}{\normalsize\bfseries}

	\fancypagestyle{firststyle}
	{
		\fancyhf{}
		\setcounter{page}{41} 
		\fancyfoot{}
		\fancyhead{}
		\fancyhead[CE]{\upshape Zeszyty Naukowe WWSI, No 28, Vol. 17, 2023, pp. \thepage-\pageref{LastPage} \newline DOI: 10.26348/znwwsi.28.\thepage}
		\fancyhead[CO]{\upshape Zeszyty Naukowe WWSI, No 28, Vol. 17, 2023, pp. \thepage-\pageref{LastPage} \newline DOI: 10.26348/znwwsi.28.\thepage}
		\fancyfoot[L,R]{\footnotesize\bfseries Manuscript received June 16, 2023}
		\fancyfoot[LE,RO]{\thepage}
		}

\title{\large \bfseries On the Correlation between Random Variables and their Principal Components} 
\author{\normalsize Zenon Gniazdowski\thanks{E-mail: zgniazdowski@wwsi.edu.pl}}
\affil{\normalsize Warsaw School of Computer Science}

\date{\vspace{-5ex}}
\providecommand{\keywords}[1]{\textbf{\textit{Keywords ---}} #1}

\begin{document}
	
	\maketitle 
	\thispagestyle{firststyle} 
	
	\linia
	
	\begin{abstract}\label{abstract}
		\noindent The article attempts to find an algebraic formula describing the correlation coefficients between random variables and the principal components representing them. 
		As a result of the analysis, starting from selected statistics relating to individual random variables, the equivalents of these statistics relating to a set of random variables were presented in the language of linear algebra, using the concepts of vector and matrix. 		
		This made it possible, in subsequent steps, to derive the expected formula. 		
		The formula found is identical to the formula used in Factor Analysis to calculate factor loadings. 
		The discussion showed that it is possible to apply this formula to optimize the number of principal components in Principal Component Analysis, as well as to optimize the number of factors in Factor Analysis.	
	\end{abstract}
	\label{keywords}
	
	\section{Introduction}
	The article \cite{Gniazdowski2017} presents the practical operation of Principal Component Analysis. The papers \cite{Gniazdowski2017} and \cite{Gniazdowski2021} analyzed the correlation relationships between random variables and the principal components representing them.  On the other hand, the article \cite{Gniazdowski2021} empirically showed that the correlation coefficients between random variables and the principal components representing them are identical to the factor loadings in Factor Analysis. This result was very interesting, also from a practical point of view. Unfortunately, there was a lack of formal confirmation that this was the correct result. This article will attempt to carry out a proof of the fact that the correlation coefficients between random variables and their principal components are the same as the factor loadings in Factor Analysis. In order to find a suitable formula, selected statistics will first be presented in the language of linear algebra. Then the algorithms of Principal Component Analysis and Factor Analysis will be presented. In the last step, the derivation of the appropriate formula will be presented.
	\section{Preliminaries}
	The preliminaries will provide definitions of basic concepts in the field of statistics that will be useful for further analysis. In particular, concepts related to a single random variable, such as mean, variance, standard deviation, will be discussed. Using these concepts, the standardization of a random variable will be discussed, as well as the linear transformation of a random variable. For two random variables, the correlation coefficient will be defined and geometrically interpreted, and the coefficient of determination will also be defined.
	Since the Hadamard product of a matrix will also be useful in this article, it will also be defined at the end of the preliminaries.
	\subsection{Mean value of the random variable}
	An n-element random variable X is considered. Its mean value $\bar{X}$ is defined as the sum of all its values divided by the number of its elements:
	\begin{equation}\label{average}
		\overline{X}=\frac{1}{n}\sum_{i=1}^{n}X_i.
	\end{equation}
	\subsection{The random component of a random variable}
	The random component of a random variable $X$ is called the difference between the variable $X$ and its mean value $\overline{X}$. If the lowercase letter $x$ describes the random component of the variable $X$, then its elements can be calculated from the following formula:
	\begin{equation}\label{losowa}
		x_i=X_i-\overline{X}.
	\end{equation}
	The resulting random component is a random variable with zero mean value. The term "random component" in the remainder of this article will often be replaced by the term "random vector". Regardless of the context, the meaning of both terms will be treated identically.
	In the remainder of this article, n-element random vectors will be considered.
	Lowercase letters will be used to denote random vectors of single random variables, e.g. variables $x$, $y$ or $p$. 
	Uppercase letters will be used to denote matrices that contain random vectors in columns, such as matrices $X$ and $P$. 
	The successive columns in the $X$ and $P$ matrices will represent successive random vectors $\{x_1,x_2,\ldots,x_n\}$ and $\{p_1,p_2,\ldots,p_n\}$, respectively. 
	\subsection{Variance of a random variable}
	A measure of the dispersion of a random variable $X$ is its variance $v$. The variance is defined as the mean value of the squares of successive components of a random vector $x$. The estimator of the variance $v$ is given by the formula:
	\begin{equation}\label{var}
		v=\frac{1}{d}\sum_{i=1}^{n}x_i^2.
	\end{equation}
	For $d=n-1$ the variance estimator $v$ is an unloaded estimator. For $d=n$ it is a loaded estimator. The loaded estimator gives an underestimate of the variance. The ratio of the estimates of the loaded variance estimator to the unloaded variance estimator is $(n-1)/n$. The limit value of this ratio is:
	\begin{equation}\label{limes}
		\lim_{n\rightarrow\infty}{\left(\frac{n-1}{n}\right)}=1.
	\end{equation}
	This means that as the value of $n$ increases, the loaded estimator follows the unloaded estimator. In view of this, it can be said that for $d=n$ the variance estimator is an asymptotically unloaded estimator \cite{Francuz2007}. For example, for $n>30$ the estimation error of the loaded estimator is less than $3.3\%$, while for $n>150$ the error is less than $0.67\%$.
	
	For the purposes of this article, it is assumed that $n$ is large enough to take $d=n$ in the formula (\ref{var}). This means that in the rest of the article the variance will always be estimated from the formula:
	\begin{equation}
		v=\frac{1}{n}\sum_{i=1}^{n}x_i^2.
	\end{equation}
	\subsection{Standard deviation of the random variable}
	The standard deviation of a random variable $X$ is defined as the square root of the variance $v$. The estimator of the standard deviation will be denoted as $s$:
	\begin{equation}\label{std}
		s=\sqrt{v}=\sqrt{\frac{1}{n}{\sum_{i=1}^{n}x_i^2}}
	\end{equation}
	\subsection{Standardized random variable}
	If a variable $X$ comes from a normal distribution with a mean $\overline{X}$ and standard deviation $s$, it can be further standardized by performing the following transformation:
	\begin{equation}
		x_i:=X_i-\overline{X}.
	\end{equation}
	After the standardization, the variable $x$ has a mean value of $\overline{x}=0$ and a standard deviation of $s=1$. If $x$ is a random vector, then its standardization involves rescaling all its components:
	\begin{equation}
		x_i:=\frac{x_i}{s}.
	\end{equation}
	\subsection{Linear transformation of a random variable}
	A random variable $X$ can be represented as the sum of its mean value $\overline{X}$ and its random component $x$:
	\begin{equation}
		X=\overline{X}+x.
	\end{equation}
	It can be seen that the linear transformation $(a+bX)$ of the random variable $X$ does not change the direction of the vector representing its random component:
	\begin{equation}\label{linear}
		a+bX=a+b\left(\overline{X}+x\right)=a+b\overline{X}+bx.
	\end{equation}
	The constant $a+b\overline{X}$ represents the mean value of the variable $X$ after a linear transformation. In turn, the random component $x$ after transformation (\ref{linear}) becomes the new random component $bx$. The vector $bx$ is parallel to the vector $x$. This means that a linear transformation of a random variable can at most change the orientation of the vector representing the random component, while it does not change the direction of this vector \cite{Gniazdowski2022}. It can be noted that the standardization of a random variable is a linear transformation, so it does not change the direction of the random vector.
	\subsection{Geometric interpretation of the correlation}\label{linTranCor}
	A measure of the relationship between two random variables is their covariance:
	\begin{equation}
		Cov_{X,Y}=E\left[\left(X-\overline{X}\right)\left(Y-\overline{Y}\right)\right].
	\end{equation}
	The covariance scaled by the standard deviations of the two random variables is called the Pearson correlation coefficient:
	\begin{equation}\label{cor1}
		r_{X,Y}=\frac{{Cov}_{X,Y}}{s_Xs_Y}=\frac{E\left[\left(X-\overline{X}\right)\left(Y-\overline{Y}\right)\right]}{\sqrt{E\left[\left(X-\overline{X}\right)^2\right]}\sqrt{E\left[\left(Y-\overline{Y}\right)^2\right]}}.
	\end{equation}
	Expression (\ref{cor1}) can be further transformed to the form:
	\begin{equation}
		r_{X,Y}=\frac{\sum_{i=1}^{n}\left[\left(X_i-\overline{X}\right)\left(Y_i-\overline{Y}\right)\right]}{\sqrt{\sum_{i=1}^{n}\left(X_i-\overline{X}\right)^2}\sqrt{\sum_{i=1}^{n}\left(Y_i-\overline{Y}\right)^2}}.
	\end{equation}
	Denoting the random components of the variables $X$ and $Y$ as $x_i=X_i-\overline{X}$ and $y_i=Y_i-\overline{Y}$, the formula for the correlation coefficient can be transformed into the form:
	\begin{equation}\label{corel}
		r_{X,Y}=\frac{\sum_{i=1}^{n}{x_iy_i}}{\sqrt{\sum_{i=1}^{n}x_i^2}\sqrt{\sum_{i=1}^{n}y_i^2}}.
	\end{equation}
	The numerator of the above formula contains the scalar product of the random vectors $x$ and $y$, while the denominator contains the product of the lengths of these vectors. The above fraction expresses the cosine of the angle between the two vectors. This means that the correlation coefficient is identical to the cosine of the angle between the random components of the random variables \cite{Gniazdowski2013}:
	\begin{equation}
		R_{X,Y}=\frac{\sum_{i=1}^{n}{x_iy_i}}{\sqrt{\sum_{i=1}^{n}x_i^2}\sqrt{\sum_{i=1}^{n}y_i^2}}=\frac{x\cdot{y}}{\|x\|\cdot\|y\|}=cos\angle(x,y).
	\end{equation}
	If two different random variables are subjected to two different linear transformations, the directions of the random components of these variables will also not change. This means that the modulus of the cosine of the angle between the two random vectors also does not change as a result of these transformations. Since the correlation coefficient is formally identical to the cosine of the angle between the vectors representing the random components of the two random variables, the linear transformation does not change the modulus of the correlation coefficient \cite{Gniazdowski2022}.
	\subsection{Coefficient of determination}
	The coefficient of determination is defined as the square of the correlation coefficient. As the square of the correlation coefficient, the coefficient of determination is a non-negative fractional number:
	\begin{equation}
		0\le R^2\le{1}.
	\end{equation}
	As a result, it can also be presented as a percentage. The coefficient of determination is a measure of the common variance of two standardized random variables. By this, it is also a good measure to describe the similarity between correlated random variables \cite{Gniazdowski2017}.
	
	Here it is important to note another feature of the linear transformation of a random variable. Subsection \ref{linTranCor} notes that the linear transformation does not change the modulus of the correlation coefficient. This also means that a linear transformation does not change the coefficient of determination.
	\subsection{Hadamard product of two matrices}
	Two matrices of identical size are considered. Each of these matrices has $m$ rows and $n$ columns. Although both matrices have identical sizes, they are not necessarily square matrices. In general, it can also be the case that $m\ne n$. The Hadamard product (or Schur product) of the two matrices $A_{m\times n}$ and $B_{m\times n}$ is a matrix $C$ of identical size to the matrices $A$ and $B$: $C_{m\times n}$. Denoting the Hadamard product operation by the asterisk symbol, the matrix $C$ resulting from the Hadamard product of matrices $A$ and $B$ can be written as:
	\begin{equation}\label{Hadam}
		C=A\ast B.
	\end{equation}
	The elements of matrix $C$ are calculated as follows \cite{Johnson2013}\cite{Liu2008}\cite{Million2007}:
	\begin{equation}\label{hadamard}
		\forall_{i,j}\;C_{ij}:=A_{ij}\cdot B_{ij}.
	\end{equation}
	It is worth noting that if in formula (\ref{Hadam}) the matrices $A$ and $B$ are identical, then there is a case of Hadamard product of a matrix by itself. In such a situation, by analogy with other known cases of products, it is obvious to talk about the Hadamard square of a matrix.
	\section{Selected statistics in the language of linear algebra}
	In the remainder of this article, using the concepts of vector and matrix, the basic concepts and operations found in statistical analysis will be presented in the language of linear algebra. First, basic statistics for single random variables will be presented. Then useful concepts and operations on sets of random variables will be defined. Also, the algorithms of Principal Component Analysis and Factor Analysis will be reminded.
	\subsection{The length of the random vector and its variance}
	A random vector $x$ is given, which has zero mean value. The variance of this random vector is given by the formula:
	\begin{equation}
		v=\frac{1}{n}\sum_{i=1}^{n}x_i^2=\frac{x\cdot x}{n}=\frac{1}{n}\left( x^Tx\right) .
	\end{equation}
	In turn, the square of the length of the vector $x$ can be calculated from the following formula:
	\begin{equation}
		\|x\|_2^2=\sum_{i=1}^{n}x_i^2=x\cdot x=x^Tx.
	\end{equation}
	The square of the length of a random vector $x$ can be related to its variance:
	\begin{equation}
		\|x\|_2^2=nv.
	\end{equation}
	As a result, the length of a random vector $x$ can be represented as a function of its variance $v$:
	\begin{equation}\label{lenVsVar}
		\|x\|_2=\sqrt{nv}.
	\end{equation}
	From here it can be seen that if a given random vector represents a standardized random variable with unit variance, then the length of this random vector is equal to the square root of the number of elements of this vector:
	\begin{equation}
		\|x\|_2=\sqrt{n}.
	\end{equation}
	\subsection{Covariance for a given matrix of random vectors}
	The covariance between two random vectors $x$ and $y$ can be represented as a function of their scalar product:
	\begin{equation}\label{covVect}
		c=\frac{1}{n}\sum_{i=1}^{n}{x_iy_i}=\frac{1}{n}\left(x\cdot y\right) =\frac{1}{n}\left( x^Ty\right).
	\end{equation}
	Instead of two random vectors $x$ and $y$, one can consider an m-element set of random vectors $\{x_1,x_2,\ldots,x_m\}$. From these $m$ vectors, a matrix $X_{n\times{m}}$ can be formed. In this matrix, the i-th column represents the i-th random vector. Generalizing formula (\ref{covVect}), a covariance matrix $C$ can be found for the matrix $X$:
	\begin{equation}
		C=\frac{1}{n}\left(X^TX\right)
	\end{equation}
	\subsection{Standardization of a matrix containing random vectors}
	Standardizing the random vectors in the columns of the $X$ matrix requires dividing all the columns of the $X$ matrix by the standard deviations of those columns.
	
	Let $V$ be a diagonal matrix containing on the diagonal the variances of the successive columns of the matrix $X$:
	\begin{equation}\label{matWar}
		V=\left[\begin{matrix}v_1&\ldots&0\\\vdots&\ddots&\vdots\\0&\ldots&v_n\\\end{matrix}\right].
	\end{equation}
	The square root of the $V$ matrix will represent the standard deviation matrix of the individual columns of the $X$ matrix:
	\begin{equation}
		\sqrt{V}=V^\frac{1}{2}.
	\end{equation}
	The algorithm required by standardization to divide the columns of the X matrix by their standard deviations can be written as follows:
	\begin{equation}
		X:=X{V^{-\frac{1}{2}}}.
	\end{equation}
	\subsection{Correlation matrix calculated for matrix $X$ containing correlated random vectors}
	Using formula (\ref{corel}), the correlation coefficient between the random vectors $x$ and $y$ can be represented as the ratio of the scalar product of the random vectors $x$ and $y$ to the product of the lengths of the random vectors $x$ and $y$.
	Since the relationship (\ref{lenVsVar}) between the length of a random vector and its variance is known, so the correlation coefficient can be represented as a function of the scalar product of two random vectors and the variance of those vectors:
	\begin{equation}\label{corOfVar}
		r_{x,y}=\frac{x\cdot y}{\sqrt{nv_x}\cdot\sqrt{nv_y}}=\frac{1}{n}\left(\frac{1}{\sqrt{v_x}}\cdot x^T y\cdot\frac{1}{\sqrt{v_y}}\right).
	\end{equation}
	The right-hand side of equation (\ref{corOfVar}) can be generalized to the case where a matrix of random vectors $X$ is given, as well as a matrix $S$ containing their standard deviations (\ref{matWar}):
	\begin{equation}
		R=\frac{1}{n}\left(V^{-\frac{1}{2}}\cdot X^T\cdot X\cdot V^{-\frac{1}{2}}\right)
	\end{equation}
	When standardized random vectors $x$ and $y$ with unit variance are considered, then equation (\ref{corOfVar}) is simplified. The correlation coefficient is equal to the n times reduced scalar product of the vectors $x$ and $y$:
	\begin{equation}
		r_{x,y}=\frac{x\cdot y}{\sqrt n\cdot\sqrt n}=\frac{1}{n}\left(x^T y\right).
	\end{equation}
	For the matrix $X$, which contains standardized random vectors, the correlation matrix $R$ takes the following form:
	\begin{equation}
		R=\frac{1}{n}\left(X^TX\right).
	\end{equation}
	The components of the correlation coefficient matrix $R$ are the elements $r_{ij}$, which are the correlation coefficients between successive random vectors $x_i$ and $x_j$:
	\begin{equation}\label{corMat}
		R=\left[\begin{matrix}1&\cdots&r_{1n}\\\vdots&\ddots&\vdots\\r_{n1}&\cdots&1\\\end{matrix}\right].
	\end{equation}
	\subsection{Matrix of coefficients of determination $D$ calculated a from the matrix of correlation coefficients $R$}
	The coefficient of determination is the square of the correlation coefficient. Therefore, for a given matrix of correlation coefficients $R$, the matrix of determination coefficients $D$ can be found as the Hadamard product of the matrix $R$ by itself. In view of this, based on formula (\ref{Hadam}), it can be said that the matrix of determination coefficients $D$ is the Hadamard square of the matrix of correlation coefficients $R$: 
	\begin{equation}
		D=R\ast R.
	\end{equation}
	Analogous to formula (\ref{hadamard}), the elements of the matrix $D$ are calculated as follows:
	\begin{equation}
		\forall_{i,j}\;D_{ij}=R_{ij}\cdot R_{ij}.
	\end{equation}
	The $D_{ij}$ element of the coefficient of determination matrix contains information about the level of common variance between the i-th variable and the j-th variable.
	\section{Application of correlation matrix in selected machine learning algorithms}
	In addition to the matrix of standardized primary variables $X$, and in addition to the statistics calculated from the matrix $X$, some machine learning algorithms (e.g. Principal Component Analysis or Factor Analysis) may require additional matrices obtained when solving the eigenproblem for the correlation matrix $R$. These include the eigenvector matrix $U$, the eigenvalue matrix (variance matrix) $\Lambda$, and the standard deviation matrix $S$. From this point of view, the algorithm for calculating these additional matrices is important.
	\subsection{Eigenproblem for correlation matrix}
	The result of solving the eigenproblem for the correlation matrix R (\ref{corMat}) is the set of eigenvalues $\{\lambda_1,\lambda_2,\ldots,\lambda_n\}$ and the corresponding set of eigenvectors $\{u_1,u_2,\ldots,u_n\}$.
	The resulting set of eigenvalues should be sorted non-increasingly.  Also note that the mutual assignment of eigenvectors to eigenvalues must not change with the sorting process. 
	
	From the eigenvalues, a diagonal matrix $\Lambda$ can be formed,  containing on the diagonal the non-increasingly sorted successive eigenvalues of $\lambda_i$:
	\begin{equation}\label{Lambda}
		\Lambda=\left[\begin{matrix}\lambda_1&\cdots&0\\\vdots&\ddots&\vdots\\0&\cdots&\lambda_n\\\end{matrix}\right].
	\end{equation}
	The $S$ diagonal matrix can also be formed from the $\Lambda$ diagonal matrix:
	\begin{equation}\label{stdMAt}
		S=\sqrt\mathrm{\Lambda}=\left[\begin{matrix}\sqrt{\lambda_1}&\cdots&0\\\vdots&\ddots&\vdots\\0&\cdots&\sqrt{\lambda_n}\\\end{matrix}\right].
	\end{equation}
	All eigenvectors are reduced to unit length. A matrix $U$ is formed from the eigenvectors. The successive columns of this matrix contain successive eigenvectors corresponding to successive eigenvalues after sorting:
	\begin{equation}\label{matU}
		U=\left[\begin{matrix}U_{11}&\cdots&U_{1n}\\\vdots&\ddots&\vdots\\U_{n1}&\cdots&U_{nn}\\\end{matrix}\right].
	\end{equation}
	\subsection{Principal components calculated based on X matrix containing standardized random variables}
	A matrix $X$ containing standardized random vectors is given. For a matrix $X$, it is possible to find its representation in the form of a principal component matrix $P$. The matrix $P$ in successive columns contains independent principal components with successively decreasing variances. Since the principal components with the smallest variances carry a negligible amount of information about the primary variables, they can be omitted. In the $P$ matrix, only those principal components will remain, which do not have too small a variance and thus carry enough information about the primary variables. The $P$ matrix will become a $P^\prime$ matrix with fewer columns. Here it can be added that the analysis of the variance of the principal components with their reduction is analogous to lossy compression of information. 
	The algorithm for finding principal components consists of the following several steps \cite{Gniazdowski2017}:
	\begin{enumerate}
		\item For a given matrix $X$, calculate the matrix of correlation coefficients $R$ (\ref{corMat}).
		\item Solve the eigenproblem for the matrix $R$ by calculating the eigenvalues and corresponding eigenvectors. Reduce the eigenvectors to unit length. 
		\item Sort the eigenvalues non-increasingly. Sorting the eigenvalues also involves reordering the eigenvectors.
		\item From successive eigenvectors create successive columns of the $U$ matrix (\ref{matU}), which describes the directions of the axes of the new coordinate system. On the directions of the axes of the new coordinate system will be projected points in the space of standardized random variables. This projection will make it possible to calculate the principal component matrix $P$. To do this, use the following transformation:
		\begin{equation}\label{PC}
			P=XU.
		\end{equation}
	\end{enumerate}
	In Principal Component Analysis, each point in the space of standardized primary variables is assigned its representation in the space with rotated axes of the coordinate system. Expression (\ref{PC}) describes how to calculate the new coordinates of points after rotation of the coordinate system axes. It can be said that before the rotation of the axes of the coordinate system, as well as after their rotation, the data points are still the same, and only their representations are different. 
	
	In practice, accepting a small loss of information, many (usually most) of the principal components can be discarded. Only those principal components that carry most of the variance information of the primary variables are retained. 
	To do this, the number of principal components $k$ should be chosen so that they represent most of the variance of the individual primary variables \cite{Gniazdowski2021}. 
	
	The reduction of principal components can be carried out in two ways:
	\begin{itemize}
		\item The first way is to keep only $k$ of the first columns of the square matrix $U_{n\times n}$. This gives a rectangular matrix $U_{n\times k}^\prime$ containing $n$ rows and $k$ columns. Symbolically, this can be written as follows:
		\begin{equation}\label{Uprime}
			U_{n\times n}\rightarrow U_{n\times k}^\prime
		\end{equation}
		Having a $U^\prime$ matrix, successive rows of the matrix with the original data should be projected onto the stored eigenvectors. This can be done using the following transformation:
		\begin{equation}
			P^\prime=XU^\prime.
		\end{equation}
		\item The second way is to find a $P$ matrix containing all principal components from the transformation (\ref{PC}), and then discard the last $n-k$ columns of this matrix, keeping only $k$ of the first columns in it. What remains after discarding the last $n-k$ columns is the k-column reduced $ P^\prime$ matrix.
	\end{itemize}
	\subsection{Factor analysis}
	In contrast to the Principal Component Method, Factor Analysis does not involve the transformation of measured primary random variables into independent principal components, but rather the creation of a linear analytical multivariate predictive model of these primary variables. The model describes the standardized primary variables as a multivariate function relative to independent standardized random variables called factors. The role of the factor model is twofold. On the one hand, using random number generators, independent standardized factors can be generated. These factors used as independent variables of the factorial model will allow the Monte Carlo method to predict the random behavior of the primary variables, taking into account their mutual correlations. On the other hand, by analyzing the factor model coefficients, it is possible to infer the similarity of the primary variables to the individual factors. Further, by giving an interpretation to the factors, one can infer the common causes of random variation in the primary variables. The Factor Analysis algorithm consists of the following steps:
	\begin{enumerate}
		\item For a given matrix $X$, calculate the correlation coefficient matrix $R$ (\ref{corMat}).
		\item For the calculated correlation matrix $R$, solve the eigenproblem by calculating the eigenvalues and their associated eigenvectors. The eigenvectors should be reduced to unit length. 
		\item Sort the eigenvalues non-increasingly. Create a diagonal matrix $\Lambda$ (\ref{Lambda}) containing the non-increasingly sorted eigenvalues of $\lambda_i$ on the diagonal.
		\item Calculate the diagonal matrix $S$ (\ref{stdMAt}).
		\item From the eigenvectors, create a matrix $U$ (\ref{matU}), which in successive columns will contain successive eigenvectors corresponding to successive sorted eigenvalues.
		\item Using the $U$ and $S$ matrices, calculate the factor loading matrix $L$:
		\begin{equation}\label{loadMat}
			L=U S=\left[\begin{matrix}L_{11}&\cdots&L_{1n}\\\vdots&\ddots&\vdots\\L_{n1}&\cdots&L_{nn}\\\end{matrix}\right].
		\end{equation}
	\end{enumerate}
	Assuming that $F=\left[f_1,\ldots,f_n\right]^T$ is a set of independent standardized random variables called factors, then the set of standardized variables $X=\left[x_1,\ldots,x_n\right]^T$ can now be represented as a linear model with respect to the set of independent factors:
	\begin{equation}
		\left[\begin{matrix}x_1\\\vdots\\x_n\\\end{matrix}\right]=\left[\begin{matrix}L_{11}&\cdots&L_{1n}\\\vdots&\ddots&\vdots\\L_{n1}&\cdots&L_{nn}\\\end{matrix}\right]\cdot\left[\begin{matrix}f_1\\\vdots\\f_n\\\end{matrix}\right].
	\end{equation}
	In practice, at the cost of a slight loss of information, most of the factor loadings can be dropped, leaving only those that will reproduce most of the variance information of the standardized X variables. To do this, the number of factors k should be chosen so that the selected factors represent most of the variance of each original variable \cite{Gniazdowski2021}. The reduction of factors can be done in two ways:
	\begin{itemize}
		\item The first way is that only the first k columns are retained from the square matrix $U_{n\times n}$ of size n. This will result in a $U_{n\times k}^\prime$ matrix of size $n\times k$ (\ref{Uprime}). Similarly, the matrix $S$ retains only the first $k$ rows and the first $k$ columns. Thus, from the initial square matrix $S_{n\times n}$, a rectangular matrix $S_{k\times k}^\prime$ is formed. The matrix $S^\prime$ has a smaller number of columns. Symbolically, this can be written as follows:
		\begin{equation}
			S_{n\times n}\rightarrow S_{k\times k}^\prime .
		\end{equation}
		Given matrices $U^\prime$ and $S^\prime$, a reduced matrix of factor loadings $L^\prime$ can be computed:
		\begin{equation}
			L^\prime=U^\prime S^\prime.
		\end{equation}
		\item The second way is that from transformation (\ref{loadMat}) the matrix $L$ is calculated, which will contain all factor loadings. 
		Then the last $n-k$ columns are discarded from the $L$ matrix, keeping only the first $k$ columns in it.
		What remains of the full $L$ matrix becomes the reduced $L^\prime$ matrix.
	\end{itemize}
	After the reduction, a k-element set $F^\prime=\left[f_1,\ldots,f_k\right]^T$ was retained, containing $k$ independent standardized random variables called factors. 	
	Now, the n-element set of standardized primary variables $X=\left[x_1,\ldots,x_n\right]^T$ can be represented as a linear model with respect to the set of $k$ independent factors:
	\begin{equation}
		X=L^\prime F^\prime.
	\end{equation}
	\section{Correlation matrix between primary variables and their principal components}
	Using formula (\ref{corOfVar}), it is possible to write down the correlation coefficient between the random vector $x$ and the vector representing the principal component $p$. The standardized vector $x$ has a unit variance. On the other hand, the vector $p$ representing the principal component has a variance equal to $\lambda$. The correlation coefficient $r_{x,p}$ can therefore be represented as follows:
	\begin{equation}\label{rxp}
		r_{x,p}=\frac{1}{n}\left(x^T p\cdot\frac{1}{\sqrt\lambda}\right).
	\end{equation}
	Using matrix $\Lambda$ (\ref{Lambda}) and $S$ (\ref{stdMAt}), equation (\ref{rxp}) can be generalized to matrix form. The matrix $R_{X,P}$ containing the correlation coefficients between the matrix X containing the standardized primary variables and the matrix P containing the principal components can be represented as follows:
	\begin{equation}\label{corXP1}
		R_{X,P}=\frac{1}{n}\left(X^TP\Lambda^{-\frac{1}{2}}\right)=\frac{1}{n}\left(X^TPS^{-1}\right).
	\end{equation}
	In turn, transforming equation (\ref{PC}), the matrix $X$ can be represented as a function of matrix $P$:
	\begin{equation}\label{xIsFofP}
		X=PU^T.
	\end{equation}
	Transposing both sides of equation (\ref{xIsFofP}), we get the formula:
	\begin{equation}\label{xTrans}
		X^T=UP^T.
	\end{equation}
	After replacing $X^T$ in (\ref{corXP1}) with the right-hand side of equation (\ref{xTrans}), we get the following formula:
	\begin{equation}\label{rXPnext}
		R_{X,P}=\frac{1}{n}UP^TPS^{-1}.
	\end{equation}
	Since the principal components are pairwise mutually independent, the product $P^TP$ is a diagonal matrix, containing on the diagonal the squares of the lengths of the vectors forming the successive principal components. Also known is relationship (\ref{lenVsVar}), which relates the variances of the principal components to the squares of their lengths.
	Therefore, the following relationship occurs:
	\begin{equation}\label{PTP}
		P^TP=n\Lambda=nSS.
	\end{equation}
	$P^TP$ in equation (\ref{rXPnext}) can be substituted by the right-hand side of equation (\ref{PTP}). After substitution and ordering, the following formula is obtained:
	\begin{equation}\label{nearFinalR}
		R_{X,P}=\frac{1}{n}n\left( US\right) \left( S S^{-1}\right).
	\end{equation}
	After simplification, the final formula defining the matrix $R_{X,P}$ will remain, which describes the correlations between the primary variables from the matrix $X$ and the principal components from the matrix $P$. This formula takes its final form:
	\begin{equation}\label{RXP}
		R_{X,P}=US.
	\end{equation}
	In the resulting matrix, the rows refer to successive primary variables, while the columns refer to successive principal components. 
	It should be noted that the result obtained is identical to the factor loadings matrix (\ref{loadMat}).
	\section{Discussion}
	The matrix product $U\cdot{S}$ has a dual interpretation. In Principal Component Analysis, it is a matrix of correlation coefficients $R_{X,P}$ (\ref{RXP}) between the primary variables $X$ and the principal components $P$. The same product in Factor Analysis is the factor loadings matrix $L$ (\ref{loadMat}). In both types of analysis, this product can have its specific applications.
	
	In Principal Component Analysis, the Hadamard square of matrix (\ref{RXP}) is the matrix of coefficients of determination between the primary variables $X$ and the principal components $P$:
	\begin{equation}\label{DXP}
		D_{X,P}=R_{X,P}\ast R_{X,P}.
	\end{equation}
	Determination coefficients $D_{X,P}$ measure the common variance between the primary variables $X$ and the principal components $P$. The $D_{ij}$ element of matrix (\ref{DXP}) reports the level of variance of the i-th primary variable represented by the j-th principal component. The $D_{X,P}$ matrix makes it possible to efficiently find the number of principal components necessary to represent most of the variance of each primary variable. In addition, thanks to the analytical formula (\ref{RXP}), there is no need to calculate multiple correlation coefficients using classical formulas of high computational complexity, analogous to what had to be done in \cite{Gniazdowski2021}.
	
	In Factor Analysis, a linear multivariate predictive model of the primary variables is created. This model describes the standardized primary variables $X$ as linear functions of independent standardized random variables called factors $F$:
	\begin{equation}
		X=L\cdot F.
	\end{equation}
	The Hadamard square of the factor loadings matrix describes the variance between successive primary variables $X$, and successive factors $F$:
	\begin{equation}\label{DXF}
		D_{X,F}=L\ast L.
	\end{equation}
	The $D_{ij}$ element of matrix (\ref{DXF}) measures the level of variance of the i-th primary variable modeled by the j-th factor. The $D_{X,F}$ matrix makes it possible to efficiently find the number of factors necessary to model most of the variance of individual primary variables.
	
	Both $D_{X,F}$ (\ref{DXF}) and $D_{X,P}$ (\ref{DXP}) matrices are important for rationalizing calculations in both Principal Component Analysis and Factor Analysis \cite{Gniazdowski2021}:
	\section{Conclusions}
	The article attempts to find an algebraic formula describing the correlation coefficients between random variables and the principal components representing them. First, starting from selected statistics relating to individual random variables, the equivalents of these statistics relating to a set of random variables are presented in the language of linear algebra. For this purpose, the concepts of vector and matrix were used. Then, in the same language of linear algebra, the corresponding analysis was successfully carried out, so that the corresponding formula was derived. As a result, formula (\ref{RXP}) was derived, which allows finding a square matrix containing the corresponding correlation coefficients between the primary variables and the principal components. The rows of this matrix refer to the subsequent primary variables. The columns of this matrix refer to the subsequent principal components.
	
	The resulting formula is identical to the formula used in Factor Analysis to calculate factor loadings. The corresponding correlation coefficient matrix or factor loadings matrix calculated with this formula can be further used in both Principal Component Analysis and Factor Analysis. In Principal Component Analysis, the Hadamard square of the calculated matrix describes the common variance between the primary variables and the principal components. In Factor Analysis, the same Hadamard square of the calculated matrix describes the variance of the individual primary variables modeled by the individual factors. It has been noted that knowledge of the respective common variances is important for rationalizing the calculation of the optimal number of principal components in Principal Component Analysis, as well as the optimal number of factors in Factor Analysis \cite{Gniazdowski2021}.
	
	\section*{Acknowledgments}
    The author would like to express his gratitude to Ms. Anna Zegarowska for reading and reviewing this article before its publication.
	\bibliography{LinAlg}\label{bibliography}
	\bibliographystyle{IEEEtran}
	
\end{document}